\documentclass[11pt]{article}

\usepackage[final]{acl}

\usepackage{times}
\usepackage{latexsym}

\usepackage[T1]{fontenc}

\usepackage[utf8]{inputenc}

\usepackage{microtype}

\usepackage{inconsolata}

\usepackage{graphicx}
\usepackage{url}
\usepackage{bm}
\usepackage{amsmath}          
\usepackage{amssymb}
\usepackage{makecell}
\usepackage{subcaption}
\usepackage{multirow}
\usepackage{xcolor}
\usepackage[dvipsnames]{xcolor}

\newcommand{\code}[1]{\texttt{#1}}

\title{Hallucination Span Detection with Input-Side Evidence Alignment}

\author{Miyu Yamada \and Yuki Arase \\
        Institute of Science Tokyo \\ \texttt{yamada.m.ee1b@m.isct.ac.jp}, \texttt{arase@c.titech.ac.jp}}

\begin{document}
\maketitle
\begin{abstract}
Hallucinations remain a major obstacle to the reliable use of large language models (LLMs) in conditional text generation. Existing methods primarily assess the factuality of an entire generated text, providing limited insight into which output spans are hallucinated or how they relate to the input. We introduce the task of hallucination span detection with input-side evidence alignment, which jointly identifies hallucinated spans and aligns output tokens with the corresponding input evidence. Our approach is based on the observation that faithful output tokens are predictable from the input, whereas hallucinated tokens are not. We therefore train an encoder-based model to predict masked output tokens from the input representation, using prediction confidence for hallucination detection while naturally producing alignments to the input. Experiments show that the proposed method effectively detects hallucinated spans and identifies meaningful input-side evidence. Human evaluation confirms the quality of the predicted alignments.
\end{abstract}

\section{Introduction}
Large language models (LLMs) have substantially advanced conditional text generation tasks, including text summarization and question answering. 
Despite these remarkable improvements, they remain prone to hallucinations, generating content that is unsupported or contradicted by the input text \cite{Article_06,niu-etal-2024-ragtruth}. 
Such hallucinations undermine the reliability of LLM-generated outputs and limit their deployment in real-world applications. 
To facilitate the verification of generated content, it is important not only to detect hallucinated spans in the output but also to identify the relevant portions of the input that support or contradict them. 
Such input--output alignment is particularly valuable for long documents, where manually tracing generated content back to its source is difficult.

\begin{figure}[t]
  \centering
  \includegraphics[width=\linewidth]{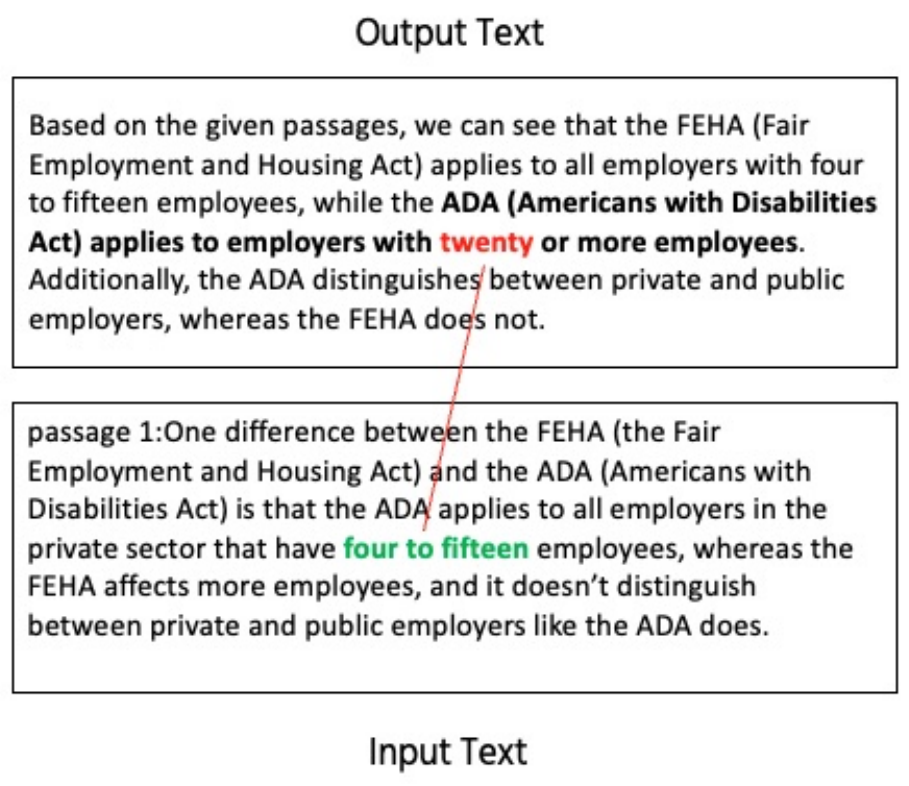}
  \caption{Example of hallucination span detection with input-side evidence alignment.
  }
  \label{fig:overview}
\end{figure}

Most existing studies have formalized hallucination detection as a binary classification problem at the sentence or document level \cite{jiang-etal-2024-large,refchecker}.
Although effective for estimating the factuality of an entire output, these approaches neither localize hallucinated spans nor identify the input evidence associated with individual output spans. This limits their usefulness for interpreting model predictions and assisting users in verifying generated content. 
To address this limitation, we introduce the task of \emph{hallucination span detection with input-side evidence alignment}, which jointly identifies hallucinated spans in generated text and aligns output spans with their corresponding evidence in the input.
Figure \ref{fig:overview} illustrates the task using text summarization as an example. 
In the output summary, the phrase of ``ADA [...] to employers with \underline{twenty} or more employees'' is hallucinated, because the source text states ``four to fifteen'' employees.
An ideal model should therefore detect the hallucinated word ``twenty'' while identifying the relevant input span of ``four to fifteen''.

Obtaining alignments between generated text and its input requires expensive manual annotation. Moreover, because LLMs evolve rapidly, constructing such annotations for every new model or domain is impractical. This motivates an approach that does not rely on manually aligned training data. 
We formulate the task under distant supervision, avoiding the need for manually aligned input-output pairs. 
Specifically, we exploit the observation that faithful output tokens are generally predictable from their supporting context in the input text, whereas hallucinated tokens are not. 
Based on this intuition, we cast hallucination span detection as an input-grounded masked token prediction problem.
Specifically, an encoder model predicts masked output tokens conditioned solely on the input representation. 
Intuitively, faithful tokens can be recovered with high confidence from the input, whereas hallucinated tokens cannot. We therefore use prediction confidence to distinguish faithful tokens from hallucinated ones.
Because each prediction is retrieved from the input representation, the framework simultaneously produces token-level evidence alignments.
Unlike recent approaches that rely on large decoder-based LLMs for hallucination detection \cite{su2025learningreasonhallucinationspan,niu-etal-2024-ragtruth,fava}, our method employs a lightweight encoder model, making inference considerably more efficient.

Experiments demonstrate that the proposed method effectively detects hallucinated spans while simultaneously identifying their supporting evidence in the input.
Human evaluation confirms the quality of the predicted alignments for both faithful and hallucinated words. 
Our code is available at \url{https://github.com/miyu-y/HalluSpan_EviAlign}.

\section{Related Work}
\subsection{Hallucination Detection}
Hallucination detection has been conventionally formulated as text-level classification, where the goal is to determine whether a generated text contains hallucination.
Representative approaches estimate consistency across multiple sampled responses \cite{manakul-etal-2023-selfcheckgpt} and LLM-based fact-checking against reference documents \cite{refchecker}.
More recently, several studies have addressed hallucination span detection. 
CoT reasoning methods \cite{akbar-etal-2024-hallumeasure} and reasoning models \cite{su2025learningreasonhallucinationspan} are employed to explain why particular spans are hallucinated. 
However, they do not necessarily provide correspondences between output spans and input evidence.
Other studies use encoder-based models to directly classify output tokens or spans \cite{lettucedetect,elchafei-abu-elkheir-2025-hallucination}. 
None of these previous studies has considered alignments of faithful and hallucinated tokens between input and generated texts.

\subsection{Monolingual Word Alignment}
Word alignment identifies correspondences between two sentences, and is therefore closely related to the alignment component of our task. 
Most monolingual word alignment methods (e.g., \citealp{lan-etal-2021-neural}) have relied on supervised learning.
However, human annotation of word alignment is expensive because it requires high expertise. 
There are a few unsupervised monolingual word alignment methods \cite{jalili-sabet-etal-2020-simalign,arase-etal-2023-unbalanced}. 
Among them, OTAlign \cite{arase-etal-2023-unbalanced} formalizes word alignment as an optimal transport problem and models null-alignments using unbalanced optimal transport.
This property is particularly relevant to our setting because hallucinated spans may not have corresponding evidence in the input. Unlike monolingual word alignment, however, our objective is not to align semantically equivalent texts but to jointly detect hallucinated spans and ground generated content in the input texts.

\section{Proposed Method}
\begin{figure}[t]
  \centering
  \includegraphics[width=\linewidth]{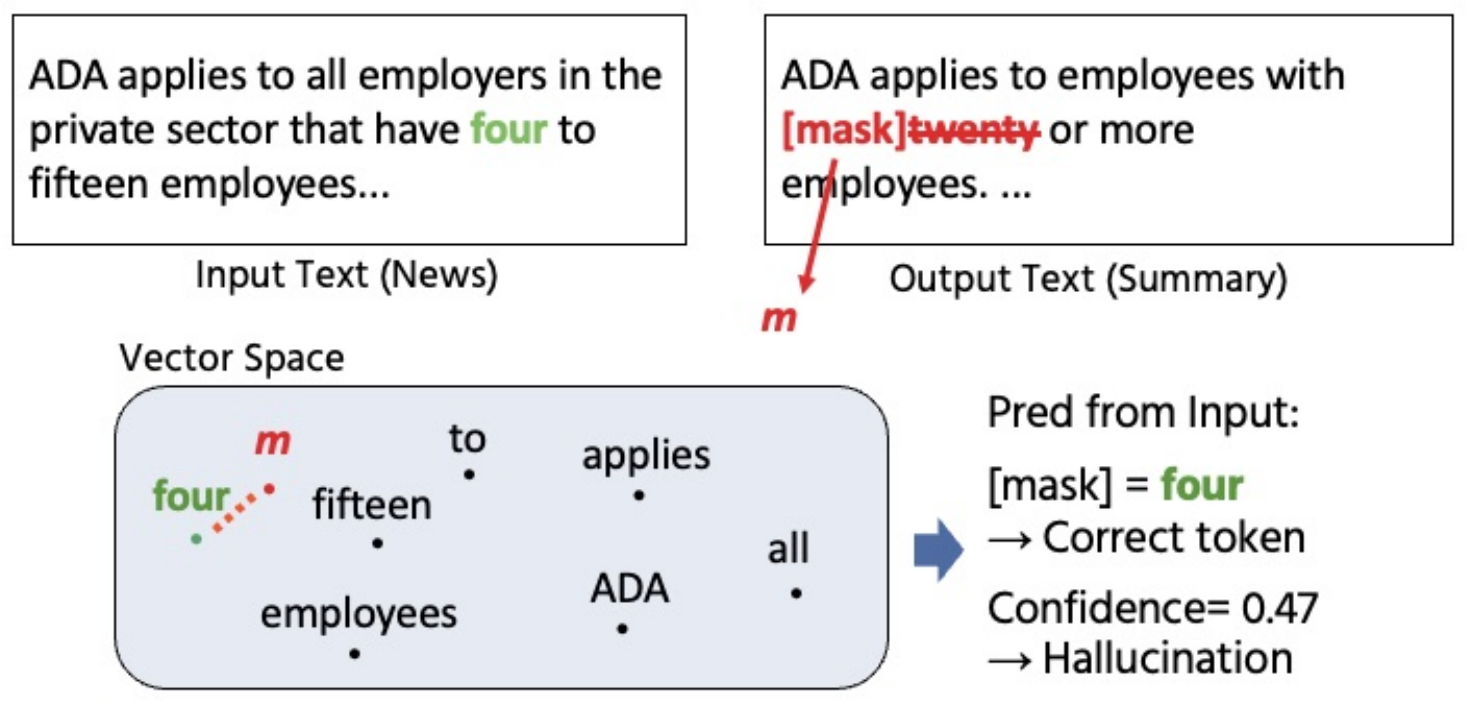}
  \caption{Proposed method}
  \label{fig:method}
\end{figure}

We assume that faithful output tokens are generally predictable from their supporting context in the input text, whereas hallucinated tokens are not. 
Based on this assumption, we formulate hallucination span detection as an input-grounded masked token prediction task.
The proposed method is inspired by the Non-Parametric Masked Language Models (NPM) \cite{min-etal-2023-nonparametric}, which recover masked tokens by matching their contextual representations with those of a reference text.
As illustrated in Figure \ref{fig:method}, the proposed method masks a token in the generated text and predicts it by retrieving the most relevant representation from the input text. 
Prediction confidence is then used to distinguish faithful tokens from hallucinated ones, while the retrieved input token provides the corresponding evidence alignment.

\subsection{Training}
\label{sec:finetuning}
For efficient training, the model is optimized using span-level masking, allowing multiple tokens to be predicted simultaneously. 

\paragraph{Span Segmentation}
We first segment output texts into semantically meaningful spans based on semantic role labeling (SRL). 
\citet{elchafei-abu-elkheir-2025-hallucination} also employed SRL-based spans for hallucination span detection. 
The below shows an example of our segmentation: 
\begin{description}
    \item[Original] Keonna Thomas was charged with attempting to travel to Syria.
    \item[Segmented]: [Keonna Thomas] [was charged] [with attempting] [to travel] [to Syria] . 
\end{description}
We extract predicates and their arguments using SRL, and segment them as spans.
When a sentence contains multiple verbs, we merge the SRL results for all predicates, and identify the finest possible spans.
Specifically, if a span identified by one predicate is further divided by another predicate, we adopt the smallest spans identified. 
Words that are not included in any predicate or argument spans, such as conjunctions, are excluded from masking.
Finally, we merge two adjacent spans of predicates to handle passive or progressive constructions. 

\paragraph{Prediction Confidence}
We then fine-tune an encoder model using span-level mask prediction.
Each masked span $s$ is replaced with two consecutive special tokens, namely, \code{<mask>}\code{<mask>}. 
The input and the masked output texts are concatenated and encoded; $\bm{m}_s \in \mathbb{R}^d$ and $\bm{m}_e  \in \mathbb{R}^d$ denote the $d$-dimensional embeddings of the former and latter \code{<mask>} token, respectively.
The embeddings of tokens of the input text, $t_1, t_2, \ldots, t_n$, are also extracted as the hidden representation of the final encoder layer:
$(\bm{d}_1,\bm{d}_2,\ldots,\bm{d}_n)$, where $\bm{d}_i \in \mathbb{R}^d$.
The prediction confidence is defined as 
\begin{equation}
  c(s,i,j)
  :=
  \mathrm{sim}(\bm{m}_s, \bm{d}_i)
  +
  \mathrm{sim}(\bm{m}_e, \bm{d}_j),
  \label{eq:score}
\end{equation}
where $i$ and $j$  are index of input tokens ($i \le j$) and $\mathrm{sim}()$ function computes cosine similarity.
The output span $s$ is then regarded as corresponding to an input span $(k, \ell)$ whose confidence is highest: $c^s_{\mathrm{max}}=\max_{(k,\ell)} c(s, k,\ell)$. 

\paragraph{Loss Function}
We design a loss function so that the model predicts input text tokens with higher confidence for faithful spans, while assigning lower confidence to hallucinated spans.
\begin{align}
  \mathcal{L} &=\frac{1}{M}\sum_{i \in M} f(s_i), \\
  f(s_i)&= 
  \begin{cases}
    \max(0, c^{s_i}_{\mathrm{max}} - \gamma_{\mathrm{h}}),
    & \text{if hallucination}, \\
    \max(0, \gamma_{\mathrm{f}} - c^{s_i}_{\mathrm{max}}),
    & \text{otherwise},
  \end{cases} \label{eq:loss}
\end{align}
where $s_i$ is the $i$-th span in the total of $M$ output spans, $\gamma_{\mathrm{h}}$ and $\gamma_{\mathrm{f}}$ are the higher and lower bounds of confidence for hallucinated and faithful spans, respectively. 
We further add a loss weight $w$ to hard examples: hallucinated spans whose confidence is higher than $\gamma_{\mathrm{f}}$ and faithful spans whose confidence are lower than $\gamma_{\mathrm{h}}$.

\subsection{Inference}
\label{sec:inference}
While we use span-level masking during training, we perform token-level prediction for inference to obtain fine-grained hallucination detection and evidence alignment.
Each output token is replaced with a single \code{<mask>} token and encoded in the same way as training. 
Finally, the input-side token with the highest confidence, computed in the same way as Equation~(\ref{eq:score}), is regarded as most strongly associated with the masked output token.


We classify each output token as hallucinated or faithful using a simple threshold $\lambda_{\mathrm{h}}$ on the maximum confidence score $c^s_{\mathrm{max}}$. 
We regard the output token as hallucinated if its maximum confidence is lower than $\lambda_{\mathrm{h}}$, and as faithful otherwise.
As a tokenizer may split a word into multiple subwords, we aggregate token-level predictions into word-level predictions.
A word is judged as hallucinated only when all of its subword tokens are classified as hallucinated.

\section{Experiment Settings}
This section describes the common experiment settings and implementation details for the automatic and human evaluations.

\subsection{Dataset}

\begin{table}[t]
\centering
\begin{tabular}{l|cc|c}
\hline
 & QA & Summary & Overall \\
\hline
Train (FT) & \makecell{$4,134$\\($8.6$\%)} & \makecell{$3,858$\\($3.0$\%)} & \makecell{$7,992$\\($5.9$\%)} \\
Dev & \makecell{$400$\\($9.0$\%)} & \makecell{$400$\\($3.3$\%)} & \makecell{$800$\\($6.1$\%)} \\
Dev ($\lambda$) & \makecell{$500$\\($9.9$\%)} & \makecell{$500$\\($3.5$\%)} & \makecell{$1,000$\\($6.7$\%)} \\
Test & \makecell{$900$\\($5.4$\%)} & \makecell{$900$\\($2.9$\%)} & \makecell{$1,800$\\($4.1$\%)} \\
\hline
\end{tabular}
\caption{Number of samples in RAGTruth (the parentheses indicate the proportion of hallucinated characters)}
\label{tab:dataset}
\end{table}

We use RAGTruth \cite{niu-etal-2024-ragtruth}, which provides outputs by $6$ different LLMs \cite{Article_02,Article_03,Article_04} on  QA, data-to-text, and news summarization tasks. 
In this study, we use the QA and news summarization tasks because the input of the data-to-text task is different from natural texts. 
In the QA task, the input consists of a passage and a question from MS MARCO \cite{Article_07}, and the output is an answer.
In the news summarization task, the input is a news article from datasets such as the CNN/Daily Mail dataset \cite{see-etal-2017-get}, and the output is its summary.
Human annotations are provided, which indicate hallucinated spans in the generated outputs. 
However, alignment between input and generated outputs is unavailable. 

\paragraph{Data Split}
Table~\ref{tab:dataset} shows the number of samples in RAGTruth with the proportion of hallucinated characters in each split. 
Because RAGTruth does not provide an official development set, we randomly extracted $800$ examples ($400$ examples for each task) from the training set (Dev).
We further randomly extracted $1,000$ examples ($500$ examples for each task) for tuning the threshold $\lambda_{\mathrm{h}}$ (Dev ($\lambda$)). 
The remaining examples were used for training.

\paragraph{Evaluation Metrics}
The official evaluation metrics of RAGTruth is character-level precision, recall, and F$_1$ score.

\subsection{Implementation of the Proposed Method}
We followed the training strategy of NPM, masking $15.0$\%  tokens of output texts.
We sampled the target mask sizes from the geometric distribution with $p=0.5$, and selected SRL-based spans whose lengths were closest to the sampled mask size. 
If a selected span contained at least one hallucinated character, it was labeled as hallucinated; otherwise, it was labeled as faithful. 
Because hallucinated spans are much fewer than faithful ones, we prioritized hallucinated spans on the mask-span sampling.
We repeated this process until the masking budget was exhausted.
As a result, the masked spans for training consisted of $35,578$ faithful and $13,269$ hallucinated spans, respectively.

For SRL-based chunking, we used the off-the-shelf SRL model released by AllenNLP \cite{SRL}.
As the base model, we employed ModernBERT-large whose parameter size is $0.4$B \cite{modernbert}.
The ModernBERT model was fine-tuned for $5$ epochs with a learning rate of $1.0$e-$5$.
We set the hyperparameters to $\gamma_{\mathrm{f}}=1.4$, $\gamma_{\mathrm{h}}=0.8$, and $w=2.0$ to maximize the F$_1$ score on Dev.
The threshold $\lambda_{\mathrm{h}}$ was determined using Dev ($\lambda$) to maximize the F$_1$ score of hallucination detection, resulting in $\lambda_{\mathrm{h}}=0.68$.


\begin{table}[t]
  \centering
  \begin{tabular}{p{0.9\linewidth}}
  \hline
  \textbf{Instruction}\\
  \hline
  Your task is to determine whether the answer contains either or both of the following two types of hallucinations:\newline$1$. conflict: instances where the answer presents direct contradiction or opposition to the passages;\newline$2$. baseless info: instances where the answer includes information which is not substantiated by or inferred from the passages.\newline Then, compile the labeled hallucinated spans into a JSON dict, with a key ``hallucination list'' and its value is a list of hallucinated spans. If there exist potential hallucinations, the output should be in the following JSON format: \{\{``hallucination list'': [hallucination span$1$, hallucination span$2$, ...]\}\}. Otherwise, leave the value as a empty list as following: \{\{``hallucination list'': []\}\}.\newline Output:\\
  \hline
\end{tabular}
\caption{Prompt for Llama-SFT}
\label{tab:prompt}
\end{table}

\subsection{Compared Methods}
We compare the proposed method with the following baseline methods.
\paragraph{Llama-SFT.}
To compare the proposed method with a larger LLM, we fine-tuned Llama-$3.1$-$8$B-Instruct \cite{grattafiori2024llama3herdmodels} for hallucination span detection.
Given an input text and an output text, the model is trained to generate hallucinated spans in the JSON format.
We adopted the same prompt as \citet{niu-etal-2024-ragtruth} (Table~\ref{tab:prompt}).
We fine-tuned the model for one epoch with a learning rate of $2.0$e-$5$ following \citet{niu-etal-2024-ragtruth}.

\paragraph{LettuceDetect.}
As the state-of-the-art in hallucination span detection models, we adopt LettuceDetect \cite{lettucedetect}. 
It fine-tunes an encoder-based LLM as a token classifier: it takes the input text and the generated output as input, and predicts whether each output token is hallucinated or not.
As the base model, we used the same ModernBERT-large as the proposed method.

\paragraph{OTAlign.}
As the baseline that can align input and output texts, we employ a word alignment method, namely, OTAlign \cite{arase-etal-2023-unbalanced}. 
It solves word alignment as the optimal transport problem. 
For each word in a generated text, we extracted aligned input words associated with transport weights. 
These aligned words are regarded as faithful while the null-aligned words are regarded as hallucinated. 
As input and generated texts in RAGTruth have significantly different lengths, we chose the unbalanced optimal transport on OTAlign for its robustness on null-alginment. 
Since RAGTruth does not provide gold-standard alignment annotations across input and generated texts, we use the unsupervised version of OTAlign. 
The hyperparameters in OTAlign were tuned using the Dev ($\lambda$) to maximize the F$_1$ score of hallucination detection. 
Namely, we set $\tau=0.02$ as the weight of the marginal relaxation term and $\lambda_{\mathrm{OT}}=0.35$ as the threshold to decide null-alignment. 

\section{Automatic Evaluation}
\begin{table}[t]
  \centering
  \resizebox{1.0\linewidth}{!}{%
  \begin{tabular}{c|ccc|ccc}
  \hline
  Method & \multicolumn{3}{c|}{QA} & \multicolumn{3}{c}{Summary}  \\
  & P & R & F$_1$ & P & R & F$_1$  \\
  \hline
  Llama-SFT& $45.0$ & $35.6$ & $39.8$ & $\textbf{64.1}$ & $37.9$ & $\textbf{47.6}$ \\
  LettuceDetect & $\textbf{66.9}$ & $62.1$ & $\textbf{64.4}$ & $60.2$ & $35.5$ & $44.6$  \\
  OTAlign & $8.9$ & $45.3$ & $14.9$ & $4.5$ & $34.8$ & $7.9$  \\
  Proposed & $40.9$ & $\textbf{82.7}$ & $54.8$ & $42.8$ & $\textbf{38.0}$ & $40.3$ \\
  \hline
  \end{tabular}
 }
  \caption{Performance of each method in hallucination span detection (P: Precision, R: Recall, F$_1$: F$_1$ score)}
  \label{tab:result}
\end{table}

We first conduct an automatic evaluation to investigate the capability of the proposed method to detect hallucinated spans.
Table~\ref{tab:result} shows the precision, recall, and F$_1$ scores of hallucination detection on QA and summarization tasks. 
In terms of F$_1$, LettuceDetect achieves the highest score on QA, while Llama-SFT performs best on summarization. 
In contrast, the proposed method achieved the best recall on both QA and summarization tasks, which is advantageous in a scenario where hallucination should not be missed. 
OTAlign performed poorly on hallucination detection. 
This is reasonable because OTAlign assumes alignment between a sentence pair, not the sets of sentences. 
In addition, largely different lengths of generated and input texts make the alignment problem extremely unbalanced, thus even unbalanced optimal transport may not be able to model it well.



\begin{table}[t]
\centering
\begin{tabular}{cp{0.75\linewidth}}
\hline
Score & Criterion \\
\hline
$3$
& Predicted input token is exactly the evidence. \\
$2$
& Predicted input token is in the sentence where the evidence is described, or in the relevant sentence. \\
$1$
& Predicted input token is irrelevant to the true evidence. \\
$0$
& Hallucination judgment is incorrect (false-positive or false-negative). \\
\hline
\end{tabular}
\caption{Criteria for manual assessment of input-side evidence alignment}
\label{tab:interpretability_criteria}
\end{table}

\begin{table*}[t!]
\centering
\begin{tabular}{lllr|rrrr|r}
\hline
Method &Top-$k$& Type & \# of words & $3$ & $2$ & $1$ & $0$ & Avg. \\
\hline
\multirow{2}{*}{OTAlign}
& Top-$1$
& Faithful & $588$ & $255$ & $57$ & $91$ & $185$ & $1.65$ \\
\cline{2-9}
& Top-$3$
& Faithful & $588$ & $295$ & $54$ & $54$ & $185$ & $1.78$ \\
\hline
\multirow{6}{*}{Proposed}
& \multirow{3}{*}{Top-$1$}
& Faithful & $588$ & $188$ & $267$ & $59$ & $74$ & $1.97$ \\
& & Baseless & $71$ & $40$ & $0$ & $19$ & $12$ & $1.96$ \\
& & Conflict & $41$ & $2$ & $3$ & $4$ & $32$ & $0.39$ \\
\cline{2-9}
& \multirow{3}{*}{Top-$3$}
& Faithful & $588$ & $268$ & $216$ & $30$ & $74$ & $2.15$ \\
& & Baseless & $71$ & $49$ & $0$ & $10$ & $12$ & $2.21$ \\
& & Conflict & $41$ & $2$ & $5$ & $2$ & $32$ & $0.44$ \\
\hline
\end{tabular}
\caption{Human evaluation results}
\label{tab:human_eval_results}
\end{table*}

\section{Human Evaluation}
\label{sec:human}
We conduct a human evaluation to manually evaluate the input-side evidence alignment.

\subsection{Settings}
We sampled $70$ output texts and randomly selected $10$ content words from each text, resulting in $700$ words for manual assessment.\footnote{We chose input texts with at most $3,000$ characters to reduce the evaluator's burden. In addition, we avoided sampling words segmented into subwords to ensure word-level assessment.}
For each sampled word, one of the authors inspected top-$k$ ($k=\{1,3\}$) prediction of input-side tokens and assessed the alignment quality based on the criteria in Table~\ref{tab:interpretability_criteria}. 
RAGTruth distinguishes hallucinations of ``baseless'' and ``conflict''; the former fabricates unsupported information, and the latter generates conflicting information. 
When the baseless hallucination does not have relevant information at all in the input, we allow such tokens to align with non-informative tokens such as punctuations and function words and assign a score $3$. 
We regard predictions with higher than score $2$ as useful as evidence to interpret the hallucination and faithful generations. 

We compare the proposed method with OTAlign to evaluate the quality of input-side evidence alignment. Because OTAlign represents hallucinated words as null-alignments rather than explicitly detecting them, we evaluate its alignment quality only on faithful output tokens. For each output token, we select the top-$k$ input tokens according to the optimal transport weights produced by OTAlign.

\subsection{Results}
Table~\ref{tab:human_eval_results} shows the distributions of words per score for top-$1$ and $3$ predictions. 

\paragraph{Input-side alignments produced by the proposed method are judged to provide useful evidence for both faithful and baseless hallucinated tokens.} 
For faithful and baseless hallucination words, their alignment to the input texts is assessed on average score around $2$, which means these input tokens are at least relevant evidence. 
On faithful words, OTAlign has more alignment to exact evidence tokens (i.e, score $3$) than our method, while it has more irrelevant alignment or prediction failures, too (score $1$ and $0$). 
This should be again due to the OTAlign's limitation on sentence-set alignment. 
In contrast, the alignments produced by the proposed method receive more score $2$ judgments. 
We attribute this tendency to the distantly supervised training objective. 
As shown in Equation~(\ref{eq:loss}), the model is optimized to confidently predict masked output tokens from the input for faithful spans while assigning low confidence to hallucinated spans. 
Because the objective does not directly supervise token-level alignments, the ability to recover the exact masked token relies on the pre-trained knowledge of ModernBERT. 
Nevertheless, the model consistently retrieves semantically relevant input tokens, which are often judged as useful evidence despite not exactly matching the evidence.

\paragraph{Conflict Hallucination is Challenging.} 
Although the occurrence of conflict hallucination is rare, the proposed method struggles to detect hallucinations of this category, i.e., the majority of cases are scored $0$. 
This is because, for conflicting hallucination cases, there should be highly relevant texts in the input. 
These relevant texts may result in closer token embeddings, and thus confuse the similarity-based confidence estimation of the proposed method. 
As future work, we will explore a richer modeling of confidence prediction by, for example, employing non-linear neural models.

\subsection{Case Study}

\begin{table*}[t!]
\centering
\resizebox{\linewidth}{!}{%
\begin{tabular}{p{0.08\linewidth}|p{0.37\linewidth}|p{0.37\linewidth}|c}
\hline
Type & Output Text & Input Text & Conf. \\
\hline
Faithful
& [...] and a commemorative Fiesta de la Flor event is planned for \textcolor{cyan}{April} $17$ and $18$.
& [...] The commemorative Fiesta de la Flor in Corpus Christi, Texas -- which celebrates her life -- is scheduled for \textcolor{Green}{April} $17$ and $18$ [...]
& $0.782$ \\
\hline
Faithful
& [...] it drew the attention of multiple law enforcement agencies including Atlanta Police Department, FBI and Federal Homeland Security. This led to her arrest on August $9$th, where a firearm and three computers were \textcolor{cyan}{found} in her home. [...] 
& [...] authorities obtained a search warrant for her residence. They took her into custody while executing that warrant. ``A firearm along with three computers was \textcolor{Green}{located} during the search,''  East Point police said [...] 
& $0.723$ \\
\hline
Faithful
& [...] As you drive through tolls, the system deducts the toll amounts from your balance until it reaches a certain \textcolor{cyan}{threshold} (\$$10$ for the TX-Tag). [...] 
& [...] They deduct the tolls from that balance till the balance goes below \textcolor{BurntOrange}{a} \textcolor{Green}{certain} \textcolor{Purple}{value} (\$$10$ for the TX-Tag) and they charge your credit card [...] 
& $0.767$ \\
\hline
Conflict
& [...] we can see that the \textbf{FEHA (Fair Employment and Housing Act) applies to all employers with four to fifteen employees}, while the \textbf{ADA (Americans with Disabilities Act) applies to employers with \textcolor{red}{twenty} or more employees.} Additionally, the ADA distinguishes between private and public employers, whereas the FEHA does not.  [...]
& [...] One difference between the FEHA (the Fair Employment and Housing Act) and the ADA (Americans with Disabilities Act) is that the ADA applies to all employers in the private sector that have \textcolor{Green}{four} to fifteen employees, whereas the FEHA affects more employees, and it doesn’t distinguish between private and public employers like the ADA does. [...] 
& $0.471$ \\
\hline
Conflict
& [...] The APhA has previously opposed the use of the term ``drug'' for chemicals used in lethal injection and has \textbf{\textcolor{red}{urged} laws} prohibiting pharmacists from participating in such cases. [...] 
& [...] This bolsters the association's previous positions to oppose the use of the term ``drug'' for chemicals used in lethal injection and to \textcolor{Green}{oppose} laws that require or prohibit pharmacists from participation in lethal injection cases. [...] 
& $0.787$ \\
\hline
\end{tabular}
}
\caption{
Examples of input-side evidence alignment by the proposed method. \textbf{Bold text} indicates the gold hallucination span.
\textcolor{Cyan}{Cyan} and \textcolor{red}{red} indicate masked \textcolor{Cyan}{faithful} and \textcolor{red}{hallucinated} output words, respectively.
\textcolor{Green}{Green}, \textcolor{BurntOrange}{orange}, and \textcolor{Purple}{purple} indicate the top-\textcolor{Green}{$1$}, - \textcolor{BurntOrange}{$2$}, and - \textcolor{Purple}{$3$} predicted input-side words, respectively. 
``Conf.'' colum shows the confidence values. 
}
\label{tab:qualitative_examples}
\end{table*}

Table~\ref{tab:qualitative_examples} provides examples of input-side evidence alignment by the proposed method, for faithful and hallucinated words. 
In the first example, the masked faithful word \textit{April} is aligned to the same word in the input text, directly indicating the supporting evidence.
In the second example, the masked faithful word \textit{found} is aligned to a paraphrased word of \textit{located}. 
The third faithful example shows that the top-$3$ predictions can recover word to phrase alignment; the output word \textit{threshold} is aligned to the correct evidence of \textit{a certain value} in the input text. 

In the first conflict hallucination example, the output describes that the regulation applies to employers with \textit{twenty} or more employees. 
This word is aligned to \textit{four} with lower confidence, which reveals the evidence that \textit{twenty} is a hallucination. 
In the second conflict example, the output states that the association has \textit{urged} laws, whereas the input states that the association \textit{opposes} the laws. 
Although the proposed method aligned these words, it failed to judge it as a hallucination due to the high confidence score, which was likely caused by highly similar surrounding contexts.


\section{Analysis}
\label{sec:type_task}

We further analyze the performance of the proposed method from the perspectives of hallucination types and generation tasks.

\subsection{Effect of Hallucination Types}
RAGTruth categorizes hallucinations into conflict and baseless types, each of them is further divided into evident and subtle hallucinations.
Evident cases are explicit errors such as generating unsupported entities or facts, whereas subtle cases involve more implicit or nuanced inconsistencies.
Table~\ref{tab:type_num} shows the numbers of hallucinated characters in each hallucination type. 

\begin{table}[t]
  \centering
      \begin{tabular}{l|cc}
      \hline
      Type & QA & Sum \\
      \hline
      Evident Conflict & $2,349$ & $5,812$ \\
      Subtle Conflict & $0$ & $339$ \\
      Evident Baseless Info & $23,441$ & $10,820$ \\
      Subtle Baseless Info & $5,545$ & $1,020$ \\
      \hline
      \end{tabular}
    \caption{Number of hallucinated characters for each hallucination type}
    \label{tab:type_num}
\end{table}

\begin{table}[t]
    \centering
      \begin{tabular}{l|cc}
      \hline
      Type & QA & Sum \\
      \hline
      Evident Conflict & $14.0$ & $16.6$ \\
      Subtle Conflict & - & $18.3$ \\
      Evident Baseless Info & $87.6$ & $50.1$ \\
      Subtle Baseless Info & $91.2$ & $37.7$ \\
      \hline
      \end{tabular}
    \caption{Recall of hallucination detection by type}
    \label{tab:type_result}
\end{table}

\begin{figure}[t!]
  \centering
  \includegraphics[width=0.8\linewidth]{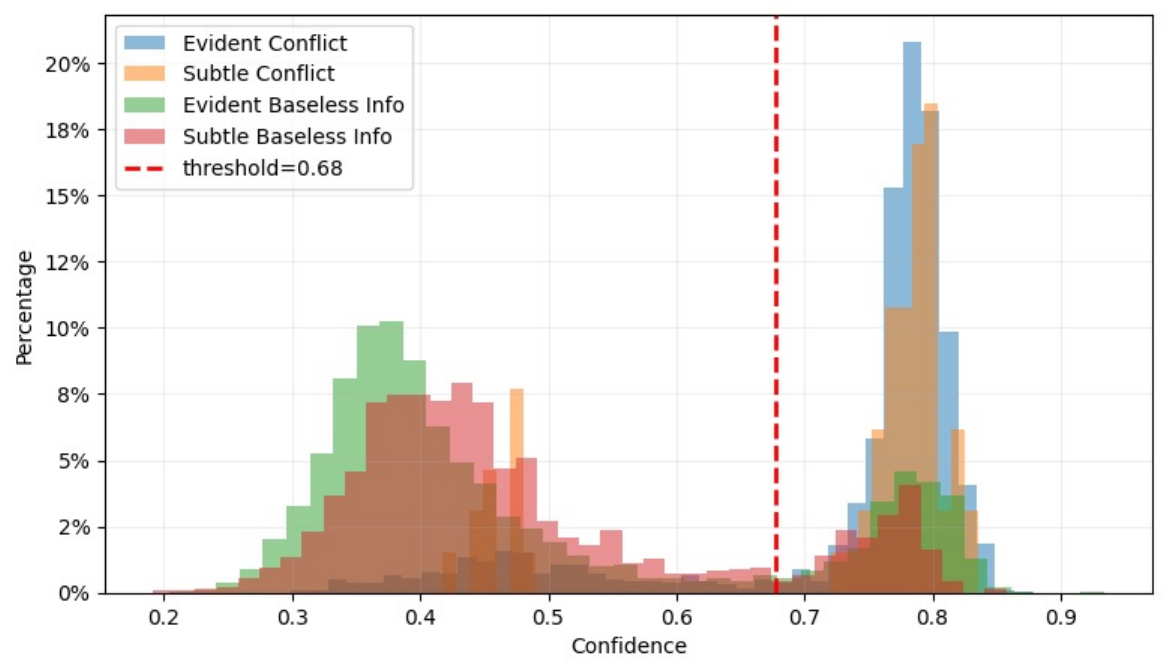}
  \caption{Distribution of maximum confidence values by hallucination type.
The red dashed line indicates the decision threshold $\lambda_{\mathrm{h}}=0.68$.}
  \label{fig:score_type}
\end{figure}

Table~\ref{tab:type_result} presents the recall of the proposed method to detect each hallucination type.\footnote{Because we do not distinguish hallucination types for detection, precision and F$_1$ are not computable.} 
Our method achieves much higher recall for baseless hallucinations than for conflict hallucinations.
The overall recall is $75.8$ and $82.9$ for evident and subtle baseless hallucinations, but only $15.8$ and $18.3$ for evident and subtle conflict hallucinations.
Figure~\ref{fig:score_type} shows the distributions of maximum confidence values for each hallucination type. 
Baseless hallucinations are concentrated in the lower value region, well below the threshold $\lambda_{\mathrm{h}}$. 
We conjecture that this is because baseless hallucinations often contain information that does not appear in the input text, thus their representations are distinctive from those of input text tokens. 
In contrast, the distribution of conflict hallucinations indicates their much higher confidence values, which makes it difficult to distinguish them from faithful tokens.

\subsection{Effects of Generation Tasks}
\begin{figure}[t!]
  \centering
    \includegraphics[width=0.8\linewidth]{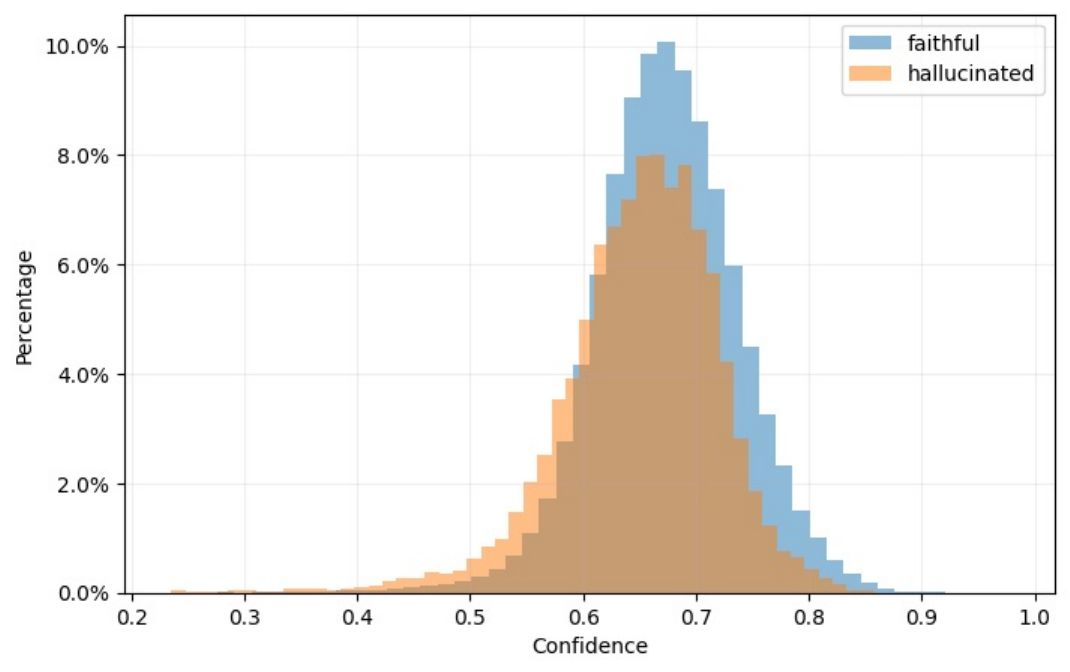}
    \caption{Original maximum confidence distributions. Blue and orange correspond to faithful and hallucinated tokens, respectively. The red dashed line represents the threshold $\lambda_{\mathrm{h}}$.}
    \label{fig:score_before}
\end{figure}

\begin{figure}[t!]
    \centering
    \includegraphics[width=0.8\linewidth]{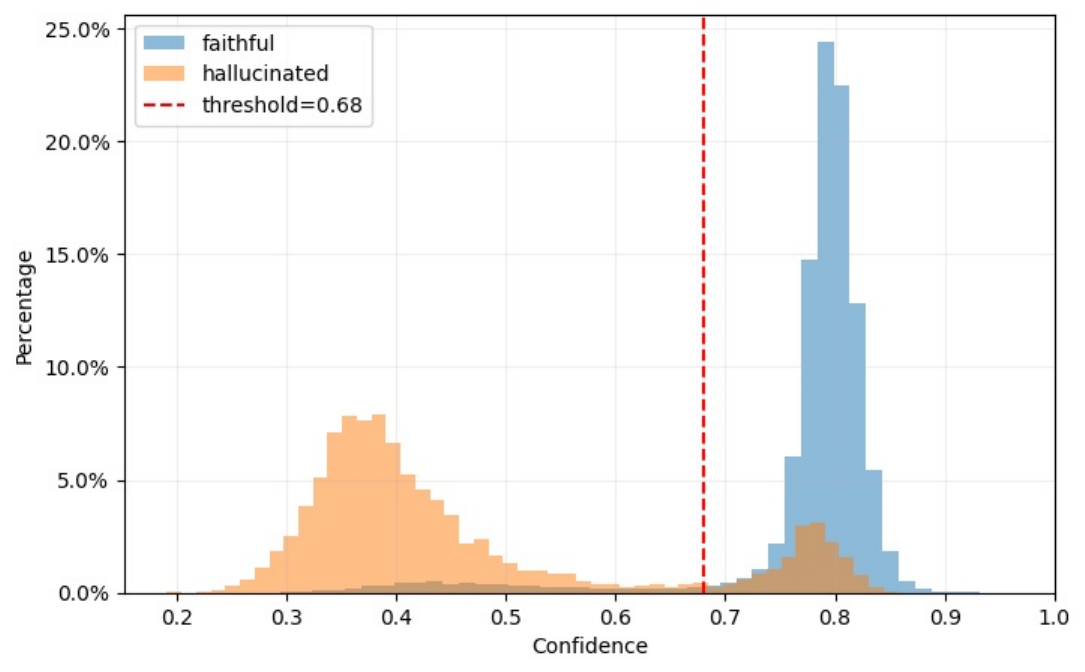}
    \caption{Maximum confidence distributions after fine-tuning on QA}
    \label{fig:score_qa}
\end{figure}

\begin{figure}[t!]
    \centering
    \includegraphics[width=0.8\linewidth]{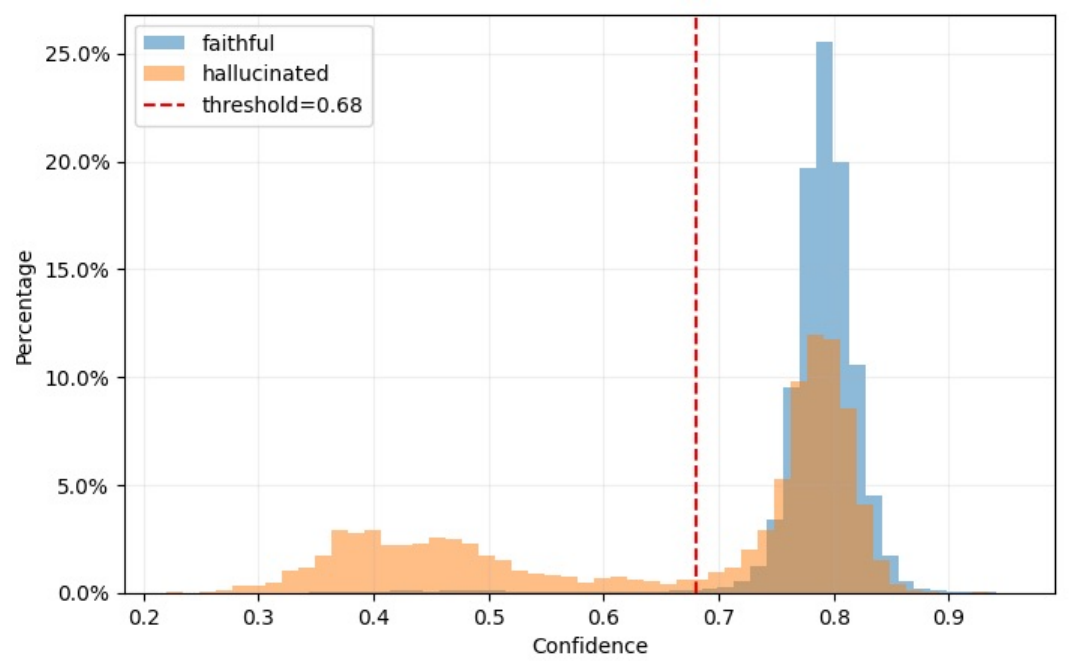}
    \caption{Maximum confidence distributions after fine-tuning on Summarization}
    \label{fig:score_sum}
\end{figure}

Figures~\ref{fig:score_before} to \ref{fig:score_sum} reveal the shift of maximum confidence distributions before and after fine-tuning. 
In the original ModernBERT model, the confidence scores, i.e., representation similarities, of faithful and hallucinated words against input texts largely overlap (Figure~\ref{fig:score_before}). 
Figures~\ref{fig:score_qa} and~\ref{fig:score_sum} indicate that the training of the proposed method effectively shifts these distributions apart to be clearly distinguishable. 
The distributions of confidence scores of hallucinated words are bimodal. 
Our observation confirms that the baseless hallucinations concentrate on the first peak with lower confidence scores, and the conflict hallucinations occupy another peak. 
This trend is more noticeable in the summarization task, as there are twice more conflict hallucinations than in the QA task while baseless hallucinations are only half of the QA task. 


\section{Conclusion}
We introduced the task of hallucination span detection with input-side evidence alignment, which aims to jointly identify hallucinated spans in generated text and align output tokens with the corresponding portions of the input. 
Although the proposed method effectively detects baseless hallucinations, improving the detection of conflict hallucinations remains an important direction for future work. 
We plan to address this limitation by developing more sophisticated confidence estimation methods for input-grounded token prediction.

\section*{Limitations}
The proposed method has several limitations. First, inference is computationally expensive because the model processes each output token independently. Specifically, the method masks one output token at a time and re-encodes both the input text and the masked output for every prediction, resulting in inference cost that grows linearly with the output length. Although the encoder model is lightweight, repeated forward passes can become expensive for long generated texts. Improving inference efficiency, for example through more efficient masking strategies or shared representations across output tokens, is an important direction for future work.

Second, our experiments are limited to question answering and news summarization using the RAGTruth benchmark. The effectiveness of the proposed framework on other conditional generation tasks, domains, languages, and LLM families remains to be investigated.

\section*{Acknowledgments}
This work was supported by JST K Program Grant Number JPMJKP$24$C$3$, Japan.
This study was carried out using the TSUBAME$4.0$ supercomputer at Institute of Science Tokyo.

\bibliography{custom}

\appendix

\end{document}